\documentclass[runningheads]{llncs}
\usepackage{graphicx}
\usepackage{amsmath}
\usepackage{amsfonts}
\usepackage{threeparttable}
\usepackage{booktabs}
\usepackage{bm}
\begin{document}
\title{INSightR-Net: Interpretable Neural Network for Regression using Similarity-based Comparisons to Prototypical Examples}
\titlerunning{INSightR-Net: Interpretable Neural Network for Regression}
%
\author{Linde S. Hesse\inst{1,2} 
Ana I. L. Namburete \inst{2,3} } 


%
\authorrunning{Hesse and Namburete}
%
\institute{ Institute of Biomedical Engineering, Department of Engineering Science \email{linde.hesse@seh.ox.ac.uk} \and
Oxford Machine Learning in Neuroimaging (OMNI) laboratory, Department of Computer Science \and Wellcome Centre for Integrative Neuroscience (WIN), Nuffield Department of Clinical Neurosciences \\University of Oxford, UK}

%
\maketitle              
\begin{abstract}
Convolutional neural networks (CNNs) have shown exceptional performance for a range of medical imaging tasks. However, conventional CNNs are not able to explain their reasoning process, therefore limiting their adoption in clinical practice. In this work, we propose an inherently interpretable CNN for regression using similarity-based comparisons (\textit{INSightR-Net}) and demonstrate our methods on the task of diabetic retinopathy grading. A prototype layer incorporated into the architecture enables visualization of the areas in the image that are most similar to learned \textit{prototypes}. The final prediction is then intuitively modeled as a mean of prototype labels, weighted by the similarities. We achieved competitive prediction performance with our \textit{INSightR-Net} compared to a ResNet baseline, showing that it is not necessary to compromise performance for interpretability. Furthermore, we quantified the quality of our explanations using sparsity and diversity, two concepts considered important for a good explanation, and demonstrated the effect of several parameters on the latent space embeddings. 

\end{abstract}
\section{Introduction}

Deep learning methods are able to achieve exceptional performance on several regression tasks in medical imaging. However, a key limitation of most of these methods is that they cannot explain their reasoning process, hampering their adoption in clinical practice \cite{markus2021role}. As clinicians are responsible for providing optimal patient care, it is crucial that they trust the model and are able to verify the model's decision. For this reason, it is important to develop models that are interpretable and can provide an explanation for their prediction. 

In interpretability research, \textit{post-hoc} methods that attempt to explain a trained black box model have been the most popular approach due to the often assumed trade-off between performance and interpretability \cite{markus2021role}. However, post-hoc explanations have shown to not always be an accurate representation of the model's decision-making process and can therefore be problematic \cite{rudin2019stop}. For example, saliency maps can resemble edge maps rather than being dependent on the trained model~\cite{adebayo2018sanity}. On the other hand, explanations obtained from an inherently interpretable model are by design an accurate representation, but developing interpretable convolutional neural network (CNN)-based architectures is a challenging task \cite{rudin2019stop,chen2019looks,li2018deep}. 

A popular interpretable CNN for the classification of natural images is \textit{ProtoPNet} \cite{chen2019looks}, which achieved comparable accuracy to several non-interpretable baselines. This method is based on learning representative examples from the training set, referred to as \textit{prototypes}, and classifies new images by computing their similarity to each of the learned prototypes. This method was later extended to breast lesion classification by incorporating part annotations \cite{barnett2021case}, and has been applied to a few other classification tasks in medical imaging~\cite{mohammadjafari2021using,singh2021these}. 

However, many medical imaging tasks are intrinsically regression problems, such as grading the severity of disease progression. While these often can be modeled as classification tasks, this ignores the linear inter-dependence of the classes, i.e. predicting a grade 4 or 2 for a true grade of 1 is penalized equally. Additionally, a human observer is likely to only fully understand an explanation if it consists of a limited number of concepts \cite{cowan2010magical}. For this reason, sparsity is considered to be important for an effective explanation \cite{rudin2019stop}. Providing an explanation for each class separately, as done in \cite{rudin2019stop}, is thus undesirable, especially for regression tasks with a large range of possible values.

In this work, we propose an Interpretable Neural Network using Similarity-based comparisons for Regression (\textit{INSightR-Net}). Our network incorporates a prototype layer \cite{chen2019looks}, providing insight into which image parts the network considers to be similar to a set of learned prototypes. The final predictions are modeled as a weighted mean of prototype labels, therefore providing an intuitive explanation for a regression task. We also propose to use a new similarity function in our prototype layer and show that this results in a sparser explanation, while maintaining the same prediction performance.

In addition to sparsity, a good explanation should also be specific to a certain sample, i.e. not all samples should have the same explanation. For this reason, we quantitatively assess both \textit{explanation sparsity} and \textit{diversity} to assess the quality of our explanations. Furthermore, in contrast to previous work \cite{chen2019looks,barnett2021case,mohammadjafari2021using,singh2021these}, we provide an analysis of the latent space representations in the prototype layer and study the effect of each of the loss components in an ablation study. We demonstrate the efficacy of our proposed \textit{INSightR-Net} on a large publicly available dataset of diabetic retinopathy grading \cite{eyepacs}.



\section{Methods}
\subsubsection{Architecture}
The network architecture consists of a ResNet-based CNN backbone (having a sigmoid as last activation), denoted by $f$, followed by a prototypical layer $g_p$ and one fully connected layer $h_{fc}$ (Fig. \ref{fig:pipeline}). Let $X \in \mathbb{R}^{w \times h \times 3}$ be a three-channel input image of size $w \times h$ with ground-truth label $y$. The feature extractor $f$ then extracts a latent representation of $X$, given by $Z \in \mathbb{R}^{w_z \times h_z \times c_z}: f(X)$, in which $c_z$ represents the number of output channels of $f$, and $w_z$ and $h_z$ are the spatial dimensions of the latent representation. By definition, both $w_z$ and $h_z$ $>1$, meaning that $Z$ can be split into latent \textit{patches} each of size $(1,1,c_z)$. Such a latent patch will be denoted by $\tilde{Z}$, and can be interpreted as the latent representation of an \textit{image part}. In the case of retinal images, $\tilde{Z}$ could for example encode optic disc information, or the presence of micro-aneurysms.

\subsubsection{Prototype Layer} 
In $g_p$, $m$ prototypes are learned, denoted by $P = \{P_j \in  \mathbb{R}^{1 \times 1 \times c_z} \; \forall j \in [1,m]\}$. Both $\tilde{Z}$ and $P_j$ can be considered as points in the same latent space and, $P_j$ can thus also be interpreted as the latent representation of an image part. The prototype layer $g_p$ aims to learn representative $P_j$'s so that for a new image $X$, a prediction can be made based on the $L_2$ distances of its latent patches, $\tilde{Z}$, to each of the prototypes in $P$. While the prototypes can initially be located anywhere in the latent space, a \textit{projection} stage in training moves each $P_j$ to the closest $\tilde{Z}$ from the training set (see \textit{Training Algorithm}).   

\begin{figure}[t]
	\centering
	\includegraphics[width=   \textwidth]{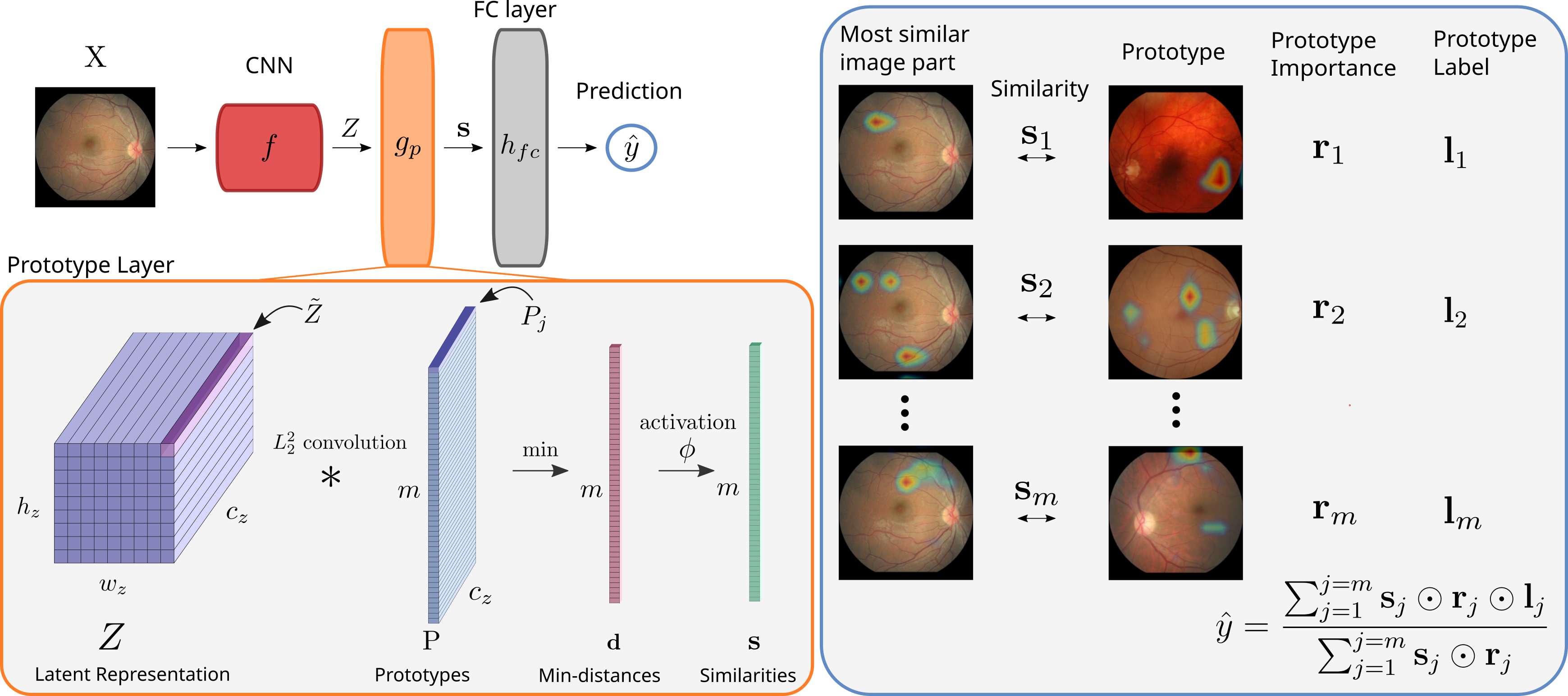}
	\caption{Overview of INSightR-Net (left) and an example prediction (right). Variables next to the boxes indicate the dimensions, unannotated sides have a dimension of 1.}
\label{fig:pipeline}
\end{figure}

As shown in Fig. \ref{fig:pipeline}, for image $X$ with latent representation $Z$, $g_p$ computes the squared $L_2$ distances between all $\tilde{Z}$ and $P_j$, followed by a min-pool operation to obtain the minimum distance of $Z$ to each of the prototypes. The distance computation is implemented using a generalized convolution with the $L_2$ norm instead of the conventional inner product \cite{nalaie2017efficient}. Effectively, this means that the learned convolutional filters of this layer represent $P$, and can be jointly optimized with the convolutional parameters of $f$. The min-distances are subsequently converted to similarities, $\mathbf{s}$, using a similarity function ($\phi(\cdot)$). Mathematically, this layer can thus be described by: $\mathbf{s} = g_{p}(Z) = \phi(\mathbf{d})$, with $\mathbf{d}$ the minimum squared $L_2$ distances between each $P_j$ and $Z$, defined by: $\mathbf{d} =  \min_{\tilde{Z} \in Z} \|\tilde{Z} - P||_2^2$.

To induce sparsity in the model predictions, we changed the logarithmic similarity function used in \cite{chen2019looks} to the following function:

\begin{equation}
    \phi(\mathbf{d}) = \frac{1}{(\mathbf{d}/d_{max}) + \epsilon}
    \label{activation}
\end{equation}

\noindent where $d_{max}$ is the maximum distance possible in latent space (exists because the last activation of $f$ is a sigmoid), and $\epsilon$ a small number to prevent division by zero. As the slope of this function increases with decreasing distance, it will amplify the differences in distances. This ultimately results in very high similarity values for some prototypes that will dominate the prediction and as such produce a sparse solution rather than one in which all prototypes contribute equally to the final prediction.

In order to make a prediction using the computed similarities, each $P_j$ is assigned a continuous label, $\mathbf{l}_j$, and should thus encode an image part that is representative for that label. In contrast to \cite{chen2019looks}, where a predefined number of prototypes per class is used, the $m$ prototypes in \textit{INSightR-Net} are continuous and span the label range of the dataset. 

\subsubsection{Fully Connected layer}
The final prediction is made by passing the similarity scores, $\mathbf{s}$, through the fully connected layer, $h_{fc}$, resulting in a single predicted value for each image, denoted by $\hat{y}$. As $\mathbf{s}$ consists of $m$ similarities, $h_{fc}$ learns $m$ weights, denoted by $\bm{\theta}_{h}$. In order to enable modeling of the final prediction as a weighted mean of prototype labels, we generate the prediction as follows: 
\begin{equation}
    \hat{y} =  \frac{\sum_{j=1}^{j=m} \mathbf{s}_j \odot \bm{\theta}_{h,j}^2 }{\sum_{j=1}^{j=m}\mathbf{s}_j \odot \bm{\theta}_{h,j}^2 / \mathbf{l}_{j}}
    \label{pred}
\end{equation}

\noindent where $\mathbf{l}_{j}$ is the label of prototype ${P_j}$, and $\odot$ denotes element-wise multiplication. Given that we can break $\bm{\theta}^2_{h}$ down into the prototype label and a weighting vector $\textbf{r}$ as $\bm{\theta}^2_{h,j} = \mathbf{r}_{j} \odot \mathbf{l}_{j}$, Eq. \ref{pred} can be rewritten into a weighted mean, with the weights consisting of the prototype similarity $\mathbf{s}$ and a weighting vector $\mathbf{r}$ as: $\mathbf{w} = \mathbf{s} \odot \mathbf{r}$. Intuitively, $\mathbf{r}$ can now be considered as a \textit{prototype importance score}. Squared weights, $\bm{\theta}_h^2$, were applied in the last layer to enforce $\mathbf{w}$ to be strictly positive, which is desired in a weighted mean computation.






\subsubsection{Training Algorithm}
\label{sect:training_alg}
We train our network with the same three-stage protocol as introduced in \cite{chen2019looks}. In the first stage, the weights of $h_{fc}$ are frozen, and both the weights of $f$ and $g_p$ are trained for a number of epochs. Next, the learned prototypes, which are the convolutional weights of $g_p$, are \textit{projected} onto the closest latent image patch. This involves replacing each prototype by the closest $\tilde{Z}$ from the training dataset, thus enabling visualization of the image corresponding to this $\tilde{Z}$. In the last stage, only $h_{fc}$ is trained to optimize prediction performance.

\subsubsection{Latent Space Regularization}
To make the distance-based predictions learn representative prototypes, we shaped our latent space using two additional loss components.  We used a cluster-based loss to pull latent patches of the image close to prototypes, creating high-density regions in the latent space containing a prototype. In contrast to pulling prototypes from the same class, as done in \cite{chen2019looks}, we replaced this expression by prototypes with a label less than $\Delta_l$ from the sample label, as the notation of the same class does not exist in a regression style prediction. Furthermore, as an image sample should have latent patches close to more than just one prototype, we replaced the minimum by a function that takes the average of the $k$ minimum values, denoted by $min_{k}$, resulting in:
\begin{equation}
    \mathcal{L}_{Clst} = \frac{1}{n} \sum_{i=1}^{n} {min_{k}(\mathbf{d}_{i,j})    \; \; \; \; \forall j: ||\mathbf{l}_j - y_i||_1 < \Delta_l}
\end{equation}


\noindent where $n$ is the number of samples in the batch and $\mathbf{d}_{i,j}$ the minimum distance between sample $i$ and prototype $j$. To facilitate prototype projection, it is desired that each prototype has at least one latent image patch nearby in latent space. As the cluster loss only attracts the $k$ closest prototypes within a certain label range, some prototypes can become outliers lacking image patches in their vicinity. For this reason, an additional prototype sample distance loss \cite{kraft2021sparrow} was included:

\begin{equation}
    \mathcal{L}_{PSD} = -\frac{1}{m} \sum_{j=1}^{m} log(1-  \min_{i \in [1,n]} (\frac{\mathbf{d}_{i,j}}{d_{max}}))
\end{equation}

The total training loss is then defined by: 
$
    \mathcal{L} = \alpha_{MSE} \mathcal{L}_{MSE} + \alpha_{Clst} \mathcal{L}_{Clst} + \alpha_{PSD} \mathcal{L}_{PSD}  
$, with $\mathcal{L}_{MSE}$ the mean squared error loss between $y$ and $\hat{y}$, and $\alpha_{(\bullet)}$ the weighting parameters of the loss components.

\section{Experimental Setup}

\subsubsection{Dataset} For this study we used the publicly available EyePACS dataset \cite{eyepacs}, which was previously used for a Kaggle challenge on diabetic retinopathy (DR) detection. The dataset consists of a large number of RGB retina images in the macula-centered field, that were labeled on the presence of DR on a scale from 0 (healthy) to 4 (most severe). The dataset consists of a training (used for training and validation) and test subset (reserved for final evaluation) and is highly unbalanced. As we considered unbalanced data to be out of scope for this study, we sampled a subset of the data ensuring label balance. Specifically, a maximum of 2443 images per grade was used from the train set, and label balance was achieved by oversampling the minority classes. In the test set we used 1206 images per grade, resulting in a total test set of 6030 images. Even though the labels in this dataset are categorical, the underlying disease progression can be considered as a regression problem. To prevent problems with negative labels, we shifted the labels to a range from 1 - 5 instead of 0 - 4, however, all results are reported without this shift. 

The retina images were preprocessed using the technique proposed by the top entry of the 2015 Kaggle challenge \cite{graham2015kaggle}. This consisted of re-scaling the retina images to have a radius of 300 pixels, subtracting the local average color and clipping with a circular mask with a radius of 270 pixels to remove boundary effects. Finally, the pre-processed images were center-cropped to 512$\times$512.

\subsubsection{Implementation}
We used a ResNet-18 as the CNN backbone \cite{he2016deep} that was pretrained as a regression task on our dataset, with the addition of one more convolutional block (Conv+ReLu+Conv+Sigmoid) to reduce the latent image size to 9$\times$9 ($w_z$ = $w_h$ = 9). The number of prototypes, $m$, was set to 50 and the depth of $Z$, $c_z$, to 128. Based on cross-validation, in the loss function $\alpha_{MSE}$ and $\alpha_{Clst}$ were set to 1 and $\alpha_{PSD}$ to 10. The $k$ in $\mathcal{L}_{Clst}$ was set to 3 and $\Delta_l$ to 0.5. 

We trained our network with two subsequent cycles of the three-step protocol, each cycle consisting of 20 epochs of the joint training stage followed by 10 epochs of last layer training. In the first 5 epochs of the first joint stage, only $g_p$ and the additional convolutional block were trained. The prototype labels were fixed at the beginning of training with equal intervals between 0.1 and 5.9 and did not change during prototype projection. The prototype label range was set slightly larger than the ground-truth labels to make sure that the network could predict values at the boundaries of the range. $\bm{\theta}_{h}^2$ was initialized to $\mathbf{l}$, effectively setting the importance of each prototype ($\mathbf{r}_j$) to 1. We performed 5-fold cross-validation using the train data and selected the best model of each fold with the validation loss. We report the average results of these models on the held-out test set.

We used the Adam optimizer with a learning rate of 1e-5 for $f$, and 1e-3 for both $g_p$ and $h_{fc}$. The batch size was set to 30 and we applied random rotation (between 0 and 360$^\circ$) and scaling (between 0.9 and 1.1) as augmentation. The same pretrained ResNet-18 backbone was used as baseline (without the additional block), adding an average pool and fully connected layer. This baseline was finetuned for 30 epochs with a learning rate of 1e-4, decaying by a factor 2 every 5 epochs. All experiments were run on a GeForce GTX 1080 GPU using Python 3.7 and Pytorch 1.7. One cross-validation fold took 2.5 hours to complete. All code is available at https://github.com/lindehesse/INSightR-Net.

\section{Results and Discussion}
To evaluate the prediction performance we used the mean absolute error (MAE) and accuracy. The prediction performance of \textit{INSightR-Net} and the ResNet-18 baseline are shown in Table \ref{tab:performance}. It is evident that the performance of \textit{INSightR-Net} is almost equal to that of the baseline, demonstrating that it is not necessary to sacrifice prediction performance to gain interpretability. 

\begin{figure}[t]
    \centering
    \includegraphics[width=0.85 \textwidth ]{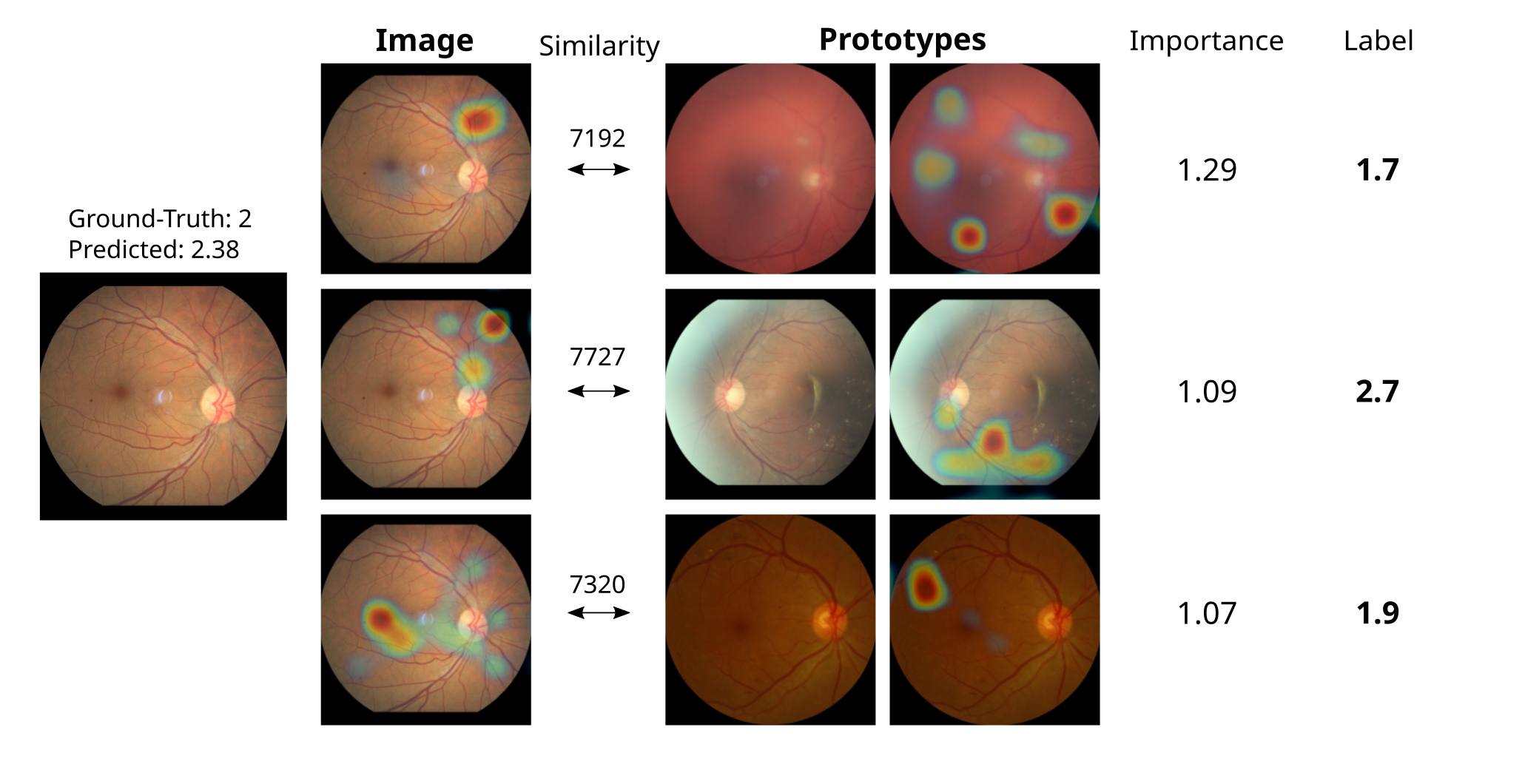}
    \caption{Example prediction showing the top-3 prototypes and their similarity scores. These top-3 prototypes constitute 33\% of the weights for the final prediction. The activation maps were upsampled from 9$\times$9, and the shown prototype labels were shifted to match the true label range (resulting in a prototype label range from -0.9 to 4.9).}
    \label{fig:predictionexample}
\end{figure}

In Fig. \ref{fig:predictionexample} the top-3 most contributing prototypes (with the highest $\mathbf{w}_j$) are shown for a representative example from the test set. The activations overlaid on top of the image sample indicate the similarity of each latent patch to the respective prototype. In a similar way, the activation maps overlaid on the prototypes represent the activation of the image that the prototype was projected on. Both of these activation maps can thus be interpreted as the regions that the model looked at while computing the similarity score. A more detailed description of the visualization of activation maps can be found in \cite{chen2019looks}.

By including a prototypical layer in our architecture, our model's decision process is thus inherently interpretable and provides not only local attention on the input image, but also similar examples (prototypes) from the training set. This contrasts with many other interpretable methods that are typically post-hoc, and provide only local attention \cite{van2022explainable}. Furthermore, our method is specifically tailored for regression and by necessity provides a single explanation for the prediction as opposed to a separate explanation for each class \cite{chen2019looks}.

\begin{table}[t]\centering
\begin{threeparttable}
\caption{Quantitative results of model predictions. Each arrow indicates the optimum of the metric. Five-class accuracy was computed from rounding regression scores to the closest integer. Shown standard deviations are across the five folds.}
\begin{tabular*}{\linewidth}{@{\extracolsep{\fill}}llllll@{}} \toprule

& MAE $\downarrow$ & Accuracy $\uparrow$  & $s_{spars}$ $\downarrow$& Diversity  $\uparrow$  \\
\midrule
ResNet-18 Baseline &  \textbf{0.59$\pm$0.004} & \textbf{0.52$\pm$0.003} &  -  &  - & \\
\midrule
(1a) with Log-Similarity \cite{chen2019looks} & 0.61$\pm$0.003 & 0.51$\pm$0.003 & 19.2$\pm$1.15 & 35.0$\pm$4.24 \\ 

(1b)  w/o $\mathcal{L}_{PSD}$ & 0.60$\pm$0.004 & 0.51$\pm$0.005 &  15.2$\pm$0.82 & 31.2$\pm$1.47  \\
(1c)  w/o $\mathcal{L}_{Clst}$ & 0.60$\pm$0.001 & \textbf{0.52$\pm$0.003} &  14.8$\pm$1.37 & \textbf{40.8$\pm$6.18}  \\

(1d) w/o $\mathcal{L}_{Clst}$ and $\mathcal{L}_{PSD}$ & 0.60$\pm$0.003 & 0.51$\pm$0.012 &  18.7$\pm$0.93 & 37.6$\pm$4.45 \\

(1e) w/o min-k in $\mathcal{L}_{Clst}$ & 0.60$\pm$0.004 & \textbf{0.52$\pm$0.008} &  \textbf{9.9$\pm$1.10}& 32.0$\pm$1.67 \\
\midrule
(1) INSightR (ours) & \textbf{0.59$\pm$0.004} & \textbf{0.52$\pm$0.005} &   14.6$\pm$0.73 & 35.8$\pm$2.64 \\
\bottomrule
\end{tabular*}
\label{tab:performance}
\end{threeparttable}
\end{table}

\begin{figure}
    \centering
    \includegraphics[width= 1\textwidth ]{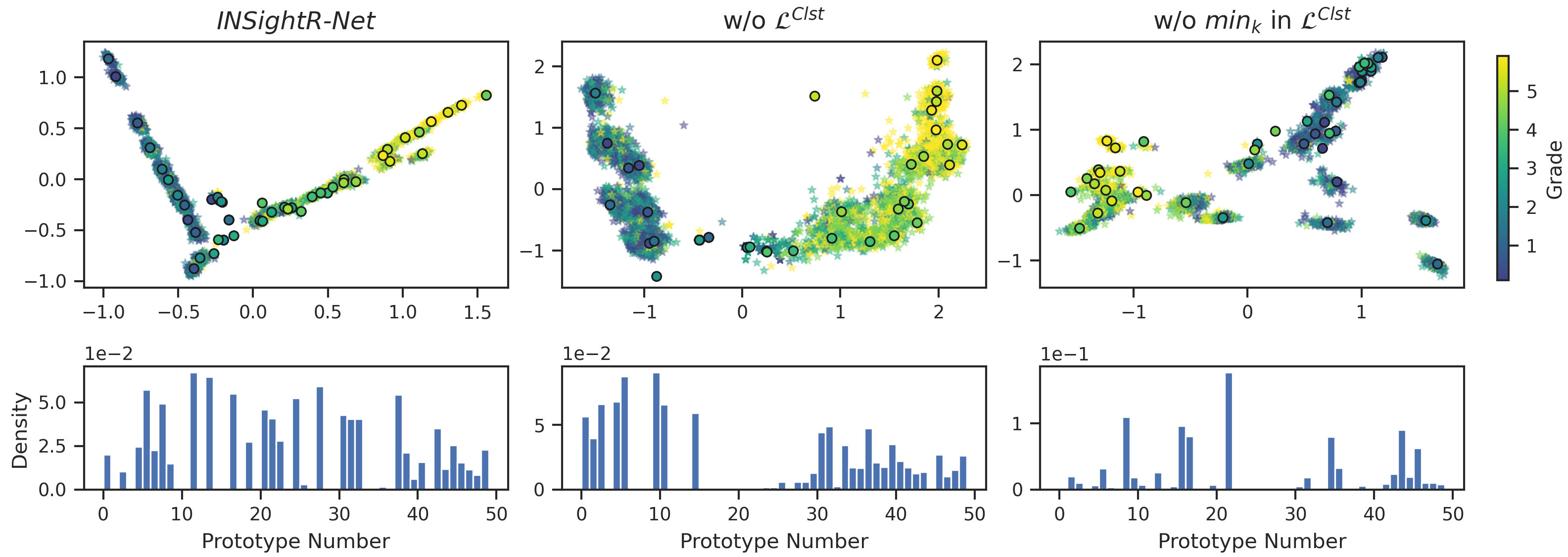}
    \caption{Top row: 2D PCA of latent space embeddings ($\tilde{Z}$) of the test set samples (stars) and learned prototypes (circles). For each sample only the 5 embeddings closest to a prototype are shown. Bottom row: Probability density histograms of the occurrence of a certain prototype in the top-5 most contributing prototypes of a test set sample.}
    \label{fig:embeddings}
\end{figure}

As it is challenging to assess the quality of the prediction from visual examples, we introduced two new metrics to quantify the quality of a prediction: explanation sparsity and model diversity. We express sparsity as the number of prototypes required to explain 80\% of the prediction (the $s_{spars}$ for which $\sum_{j=1}^{j=s_{spars}}\mathbf{w}_j = 0.8 \cdot \sum_{j=1}^{j=m} \mathbf{w}_j$). In this definition, a lower $s_{spars}$ results in a sparser solution. While sparsity is a desired quality, a low sparsity could also be a result of the same few prototypes being used for each explanation. For this reason, we also quantify model diversity, which we define by the number of prototypes that are in the top-5 contributing prototypes for at least 1\% of the test set images. Ideally, the explanation of each sample is different (high \textit{diversity}) and can mostly be captured by a minimal number of prototypes (low $s_{spars}$).

It can be seen in Table \ref{tab:performance} that replacing the log-activation as the similarity function improves sparsity ($p<0.001$ for all folds using a Wilcoxon signed-rank test) while preserving model diversity. This confirms our hypothesis that the new similarity function increases the sparsity of the explanation. Furthermore, this similarity function also increases prediction performance by a small amount.

\subsubsection{Ablation Studies}
The effect of removing loss components is shown in Table \ref{tab:performance}. It can be seen that removing $\mathcal{L}_{PSD}$ results in worse $s_{spars}$ and model diversity, clearly demonstrating the need for this component. However, removing $\mathcal{L}_{clust}$ shows a less clear effect, improving model diversity while maintaining a similar $s_{spars}$. To demonstrate the effect of this component in more detail, we visualized their latent representations in Fig. \ref{fig:embeddings}. It is evident that without $\mathcal{L}_{clust}$, the samples are spread out in latent space with prototypes scattered in these representations. On the other hand, with $\mathcal{L}_{clust}$, the samples are clustered around prototypes, thus encouraging each prototype to display a representative concept. 

Removing the $min_k$ averaging (replacing it with a minimum) in $\mathcal{L}_{Clust}$ reduces $s_{spars}$ as well as model diversity. It can be seen in Fig. \ref{fig:embeddings} that the improved sparsity is a direct result of several prototypes barely contributing to the model predictions. From the histograms it can be observed that while most prototypes are being used in the first two panels, many prototypes are almost unused when $min_k$ is removed from $\mathcal{L}_{Clust}$, demonstrating the balance between explanation sparsity and diversity.


\subsubsection{Continuous Labels}
The results presented in this section were on the (categorical) labels. To show that our method also works for real-valued labels, we transformed the labels in the train set to a continuous distribution by re-assigning each image a label from the uniform distribution $\sim U(c_l-0.5, c_l+0.5)$, with $c_l$ the categorical class label. These results are given in the supplementary material. 

\section{Conclusion}
In this work we showed that \textit{INSightR-Net} is able to provide an intuitive explanation for a regression task while maintaining baseline prediction performance. We quantified the quality of our explanations using sparsity and diversity, and demonstrated that a new similarity function and adjustments to the loss components contributed to an improved explanation. We demonstrated \textit{INSightR-Net} on a DR dataset but it can be applied to any feedforward regression CNN, offering a novel method to discover the reasoning process of the model.

\subsubsection*{Acknowledgments}
LH acknowledges the support of the UK Engineering and Physical Sciences Research Council (EPSRC) Doctoral Training Award. AN is grateful for support from the UK Royal Academy of Engineering under its Engineering for Development scheme.

\bibliographystyle{splncs04}
\bibliography{mybib}

\end{document}